\documentclass[journal]{IEEEtran}

\usepackage{tabularx}

\usepackage{comment}
\usepackage{graphicx}
\usepackage{amsmath}
\usepackage{amssymb}
\usepackage{comment}
\usepackage{multirow}
\usepackage{hyperref}
\hypersetup{hidelinks}

\usepackage{booktabs}
\usepackage{array, caption, threeparttable}
\usepackage{caption}
\usepackage{algorithm}
\usepackage{listings}
\usepackage{color}
\usepackage[dvipsnames, table]{xcolor}
\usepackage{mathtools}

\usepackage{todonotes}
\usepackage{gensymb}

\usepackage{makecell}

\usepackage{xspace}
\definecolor{baselinecolor}{gray}{0.8}

\usepackage{gensymb}
\usepackage{pifont}

\usepackage{amssymb}
\makeatletter

\newcommand{\Rmnum}[1]{\expandafter\@slowromancap\romannumeral #1@}

\usepackage{bbm}

\usepackage{subcaption}
\usepackage{algorithm}
\usepackage[table]{xcolor}
\usepackage{algorithm}
\usepackage{algorithmic}
\usepackage{xcolor}

\definecolor{b}{rgb}{0,0,0}

\begin{document}

\title{$\Delta$VLA: Prior-Guided Vision-Language-Action Models via World Knowledge Variation}

\author{Yijie Zhu,
         Jie He,
        Rui Shao$^\dagger$,~\IEEEmembership{Member,~IEEE},
        Kaishen Yuan,
        Tao Tan,~\IEEEmembership{Member,~IEEE},
        Xiaochen Yuan,~\IEEEmembership{Senior Member, IEEE},
        Zitong Yu$^\dagger$,~\IEEEmembership{Senior Member,~IEEE}
        \\

\thanks{$^\dagger$ Corresponding author: Rui Shao (email: shaorui@hit.edu.cn) and Zitong Yu (email: yuzitong@gbu.edu.cn).}

\thanks{Yijie Zhu is with Harbin Institute of Technology, Shenzhen, Shenzhen 518055, China, and Great Bay University, Dongguan 523000, China.}

\thanks{Jie He, Rui Shao are with Harbin Institute of Technology, Shenzhen, Shenzhen 518055, China.}
\thanks{Kaishen Yuan is with the Information Hub, The Hong Kong University of Science and Technology (Guangzhou), Guangzhou 511400, China.}
\thanks{Tao Tan, Xiaochen Yuan are with Macao Polytechnic University, Macao 999078, China.}
\thanks{Zitong Yu is with Great Bay University, Dongguan 523000, China.}
}


\maketitle

\begin{abstract}
Recent vision-language-action (VLA) models have significantly advanced robotic manipulation by unifying perception, reasoning, and control. To achieve such integration, recent studies adopt a predictive paradigm that models future visual states or world knowledge to guide action generation.
However, these models emphasize forecasting outcomes rather than reasoning about the underlying process of change, which is essential for determining how to act.
To address this, we propose \textbf{$\boldsymbol{\Delta}$VLA}, a prior-guided framework that models world-knowledge variations relative to an explicit current-world knowledge prior for action generation, rather than regressing absolute future world states. Specifically,
\textbf{1)} to construct the current world knowledge prior, we propose the \textbf{Prior-Guided World Knowledge Extractor (PWKE)}. It extracts manipulable regions, spatial relations, and semantic cues from the visual input, guided by auxiliary heads and prior pseudo labels, thus reducing redundancy.
\textbf{2)} Building upon this, to represent how world knowledge evolves under actions, we introduce the \textbf{Latent World Variation Quantization (LWVQ)}. It learns a discrete latent space via a VQ-VAE objective to encode world knowledge variations, shifting prediction from full modalities to compact latent. \textbf{3)} Moreover, to mitigate interference during variation modeling, we design the \textbf{Conditional Variation Attention (CV-Atten)}, which promotes disentangled learning and preserves the independence of knowledge representations.
Extensive experiments on both simulated benchmarks and real-world robotic tasks demonstrate \textbf{$\boldsymbol{\Delta}$VLA} achieves state-of-the-art performance while improving efficiency. Code and real-world execution videos are available at \url{https://github.com/JiuTian-VL/DeltaVLA}.
\end{abstract}


\IEEEpeerreviewmaketitle
\section{Introduction}
\label{sec:intro}
\begin{figure}
    \centering
    \includegraphics[width=0.98\linewidth]{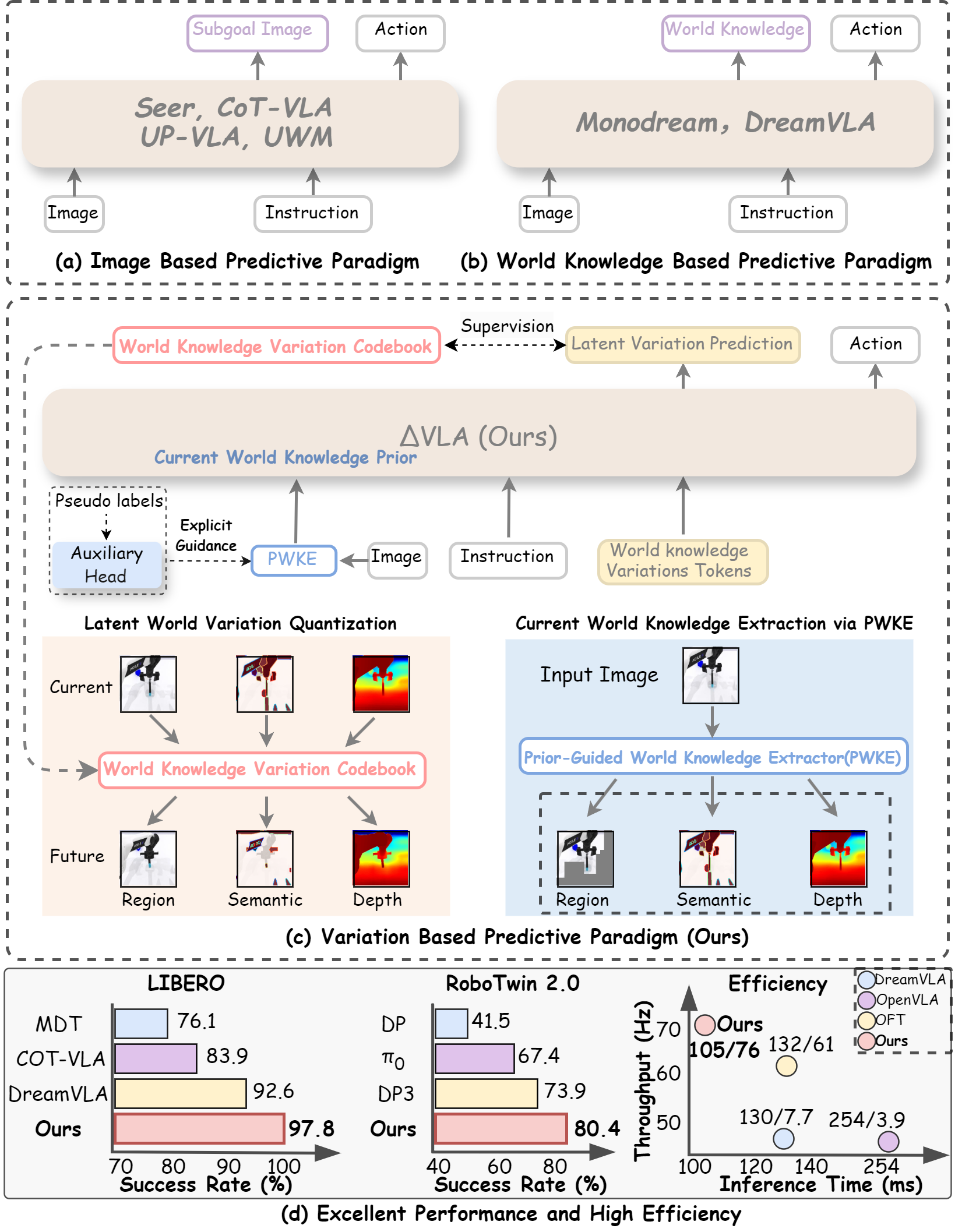}
    \vspace{-5pt}
    \caption{\textbf{Comparison with Previous Paradigms.} 
(a) Image-based paradigms predict subgoal images for action generation. (b) World knowledge-based paradigms replace pixel-level prediction with future world states. (c) \textbf{$\boldsymbol{\Delta}$VLA} constructs an explicit current-world knowledge prior via \textbf{PWKE} and models discrete world-knowledge variations via \textbf{LWVQ}, enabling prior-grounded reasoning about how the world should change under actions. (d) Comparison of performance and efficiency with previous methods.}

    \label{fig:intro}
    \vspace{-13pt}
\end{figure}
Research on Vision-Language-Action (VLA) models~\cite{chatvla, li2025lion, pi0, dexgraspvla, li2025semanticvla, li2025semanticvla, li2025cogvla, shao2025large} has advanced rapidly, propelled by large-scale pre-trained Vision-Language Models (VLMs)~\cite{improved, liu2025llm, visual, yuan2025coemogen, zhou2025hiconagent, zhang2025falcon, lyu2025puma, lyu2026personalalign, li2024optimus} that provide rich multimodal representations. Building upon these foundations, recent studies~\cite{zhen20243d,seer,zhang2025up,cot} have incorporated future knowledge prediction into VLA to unify perception, reasoning, and control for coherent and generalizable robotic behavior.
As shown in Fig.~\hyperref[fig:intro]{1(a)}, some methods~\cite{zhu2025unified, zhang2025reinbot, yang2025gripper} directly predict future images as intermediate representations for action generation. Building upon this, approaches like~\cite{wang2025monodream, zhang2025dreamvla} further replace pixel-level prediction with comprehensive world knowledge modeling, enabling richer reasoning and stronger generalization, as shown in Fig.~\hyperref[fig:intro]{1(b)}.

Despite these advances, most existing approaches still rely on a paradigm that directly predicts future world knowledge. This formulation neglects that the quality of an action is determined by the variation it induces~\cite{bruce2024genie, chi2025mind} rather than the absolute future state. 
Indeed, modeling variation has long been a standard technique in many areas~\cite{hafner2019dream, hafner2019learning, he2016deep}, as emphasizing differences can stabilize prediction and highlight transitions.
However, directly transplanting variation prediction into VLA remains insufficient for two VLA-specific reasons:
\textbf{i) Missing a causal anchor in the present.} 
Without an explicit current world-knowledge prior, the model lacks a grounded reference for deciding what should change versus remain invariant, which makes relative change prediction ill-posed and can lead to prior-free imagination and ungrounded attribution of changes to the correct entities.
\textbf{ii) Continuous deltas yield unstable conditioning.} Even with a grounded prior, unconstrained continuous variations are often highly scene- and instruction-dependent, making them ill-suited as a compact and consistent conditioning interface for policy learning, and thus motivating a discrete variation representation.
Taken together, these limitations encourage the model to fall back on full future reconstruction, favoring visual plausibility over actionable causality and learning what the world may look like rather than how it should change to satisfy the instruction. This often yields visually coherent yet behaviorally ambiguous outcomes, where fine-grained, control-critical variations are under-emphasized.

To address these limitations, we propose $\boldsymbol{\Delta}$VLA, a prior-guided framework that models world-knowledge variations relative to the current-world knowledge prior for action generation, rather than regressing absolute future world states.
As depicted in Fig.~\hyperref[fig:intro]{1(c)}, it first builds explicit current world knowledge from visual inputs and then predicts its variation to model how the environment evolves under actions. Specifically, \textbf{1)} to construct the current world knowledge prior, we propose the \textbf{Prior-Guided World Knowledge Extractor (PWKE)}. It exploits encoder specialization by leveraging the complementary strengths of SigLIP~\cite{siglip} for semantic understanding, and DINOv2~\cite{dinov2} for spatial geometry.  By utilizing auxiliary heads and pseudo labels, it explicitly supervises the extraction of three key components of world knowledge: manipulable regions from both encoders, depth cues from DINOv2, and semantic information from SigLIP.
 By integrating these cues, PWKE forms a unified current-world knowledge prior for subsequent variation modeling and alleviates redundant visual information. \textbf{2)} Building upon this prior, to represent how world knowledge evolves under actions, we introduce the \textbf{Latent World Variation Quantization (LWVQ)}. It learns a discrete latent space in an unsupervised manner via a VQ-VAE objective~\cite{van2017neural}, which discretizes continuous, scene-dependent variations into a compact set of variation tokens. During training, the model is optimized to predict these discrete variation tokens rather than reconstructing full visual modalities, providing a more stable conditioning interface for action generation. This shifts learning from full-modality prediction to compact latent reasoning, improving both efficiency and consistency in action learning. \textbf{3)} Moreover, to mitigate interference during variation modeling, we propose the \textbf{Conditional Variation Attention (CV-Atten)}.
It adopts a structured attention-masking mechanism that conditions each variation token on its corresponding world-knowledge prior while suppressing attention to irrelevant modalities.
This design enforces independence among semantic, depth, and regional variations, reducing cross-stream interference and promoting disentangled variation learning.

Compared with previous paradigms, $\boldsymbol{\Delta}$VLA offers \textbf{three key advantages:}
\textbf{i) Explicit prior grounding.} It explicitly extracts current world knowledge from observations, providing a causal anchor for reasoning while reducing prior-free imagination and redundant perception.
\textbf{ii) Prior-grounded variation reasoning.} It focuses on task-relevant changes relative to the present, rather than reconstructing absolute futures, thereby encouraging the model to capture what should change versus remain invariant under actions.
\textbf{iii) Discrete, policy-usable variation interface.} It represents variations as a discrete and transferable conditioning interface for action generation, improving learning efficiency and generalization.


We conduct comprehensive evaluations of \textbf{$\boldsymbol{\Delta}$VLA} on the LIBERO~\cite{libero} and RoboTwin~2.0~\cite{chen2025robotwin} simulation benchmarks, as well as real-world robotic manipulation tasks. As shown in Fig.~\hyperref[fig:intro]{1(d)}, the results show that our model consistently achieves state-of-the-art success rates while maintaining superior efficiency across diverse manipulation tasks. In summary, our main contributions are as follows:
\begin{itemize}
  \item We propose \textbf{$\boldsymbol{\Delta}$VLA}, a prior-guided VLA framework that represents discrete world-knowledge variations conditioned on an explicit current-world knowledge prior for action generation.
   \item We introduce \textbf{PWKE} to construct an explicit current-world knowledge prior as a causal anchor, and \textbf{LWVQ} to represent its variations in a discrete latent form for consistent policy conditioning.
    \item We develop \textbf{CV-Atten}, a structured attention mechanism that conditions each variation on its corresponding prior, ensuring interference-free variation learning.
    \item Extensive experiments on both simulation and real-world robotic tasks demonstrate the superior performance and efficiency of our \textbf{$\boldsymbol{\Delta}$VLA}.
\end{itemize}
\section{Related Work}
\label{Related Work}
 \subsection{Vision--Language--Action Models}
Early studies on Vision–Language–Action (VLA) models~\cite{shridhar2022cliport, reed2022generalist, zhang2024navid, zhang2024uni, zhu2025h, li2025cogvla, li2025semanticvla} established the foundation for integrating visual perception, language understanding, and control.
These methods leveraged pretrained vision–language features to guide task-conditioned policies for basic manipulation and decision-making. 
By leveraging the representational strengths of Vision–Language Models (VLMs) ~\cite{shen2024mome, shao2023detecting, shao2024detecting, shao2019multi, zhu2025emosym,  zhu2025uniemo, lyu2025puma}, VLA has rapidly evolved into a key paradigm for unifying perception and control in robot learning.  Octo~\cite{octo} introduced a large and diverse multi-robot dataset to facilitate multitask training. 
The RT series~\cite{Rt-1, Rt-2, Rt-h} pioneered the use of action tokenization, enabling scalable transfer from web-scale data to real robotic control. Building on multi-robot and semantic supervision, the $\pi$ series~\cite{pi0, pi05} developed a heterogeneous co-training approach that promotes transferable representations.
In parallel, OpenVLA~\cite{openvla} has been released as a 7B open-source VLA model pre-trained on approximately 970k real-world robot demonstrations, achieving strong generalist manipulation performance.
Meanwhile, motivated by the ability of diffusion models to capture multi-modal action distributions, recent works~\cite{hou2025dita, ke20243d, ze20243d} generate actions by sampling from noise conditioned on observations, task instructions, and robot priors using diffusion-based architectures.

 \subsection{Predictive World Modeling for Robotics}
Models that directly map observations and instructions to actions lack explicit reasoning processes. Thus, recent studies~\cite{seer, zhu2025unified, nasiriany2024pivot, zhen20243d} extend VLAs toward predictive paradigms to incorporate intermediate reasoning.
Some methods~\cite{hu2024video, zhang2024pivot, gu2023rt, nasiriany2024pivot} employ auxiliary generative models to synthesize future or goal images for subsequent action prediction, which increases inference latency and computation. Thus, to improve efficiency, later works~\cite{zhu2025unified, zhang2025reinbot, yang2025gripper, zhu2025h} unify future prediction and action modeling in an end-to-end framework. Specifically, UP-VLA~\cite{zhang2025up} revisits the training paradigm by jointly optimizing multi-modal understanding and future prediction objectives, thereby complementing high-level semantic reasoning with low-level spatial sensitivity that is critical for robotic control. CoT-VLA~\cite{cot} further injects explicit visual chain-of-thought reasoning by autoregressively predicting future frames as intermediate goals before generating short action sequences.
Building upon this idea, approaches such as~\cite{wang2025monodream,zhang2025dreamvla} further replace pixel-level forecasting with comprehensive world-knowledge modeling, enabling richer reasoning.
However, these methods are still largely supervised by absolute future states, which can bias learning toward reconstructing what the future looks like rather than isolating action-induced changes that matter for control, and they often lack an explicit present-time anchor. In contrast, our work constructs a current-world knowledge prior as a causal reference and models world-knowledge variations relative to this prior in a discrete, policy-usable form, providing a compact and stable conditioning interface for action generation.
\subsection{Variation Modeling in Dynamics.}
Variation modeling, also referred to as difference or residual modeling, is a long-standing idea in dynamics learning and model-based reinforcement learning~\cite{hafner2019dream, hafner2019learning, cui2024dynamo, schmidt2023learning, shao2017deep}.
PlaNet~\cite{hafner2019learning} learns a latent dynamics model from pixel observations and performs fast online planning in latent space for action selection. Related latent world-model approaches further scale this paradigm by learning compact transition representations for imagination-based control, such as Dreamer and its variants~\cite{hafner2019dream, hafner2023mastering, hafner2020mastering}.
Beyond latent world models, probabilistic ensemble dynamics with trajectory sampling~\cite{chua2018deep} and subsequent model-based RL methods similarly rely on learned transition models to enable planning and policy improvement.
 While residual/difference modeling is widely used in dynamics learning for transition prediction and planning, directly transferring it to VLA is insufficient. VLA requires present-anchored change representations: without an explicit current-world knowledge prior, variation prediction lacks a grounded reference for what should change versus remain invariant under an instruction. Moreover, continuous deltas are often highly scene- and instruction-dependent, making them an unstable conditioning interface for action generation. In contrast, we construct a current-world knowledge prior (PWKE) and model discrete world-knowledge variations relative to it (LWVQ) for stable conditioning, with CV-Atten reducing cross-stream interference during variation learning.

\section{Preliminaries and Problem Formulation}
\noindent\textbf{Previous Predictive World Modeling.} 
At each timestep $t$, given the current context $\mathbf{X} = \{\mathit{O}_t, \mathit{I}\}$ 
that includes the visual observation and task instruction, 
the model first predicts the future world knowledge $\mathit{W}_{t+n}$ 
(e.g., semantic and geometric properties). 
It then produces the corresponding $n$-step action sequence 
$\mathit{A}_{t:t+n-1} \in \mathbb{R}^{n \times d}$ 
conditioned on this world knowledge, 
where $d$ represents the dimension of each atomic action 
(e.g., $d = 7$ for 3-DoF translation $\Delta T$, 3-DoF rotation $\Delta R$, and binary gripper control). 

\noindent\textbf{Problem Formulation.} 
This predictive paradigm focuses on regressing the absolute future world knowledge $\mathit{W}_{t+n}$.
In VLA, however, action generation requires reasoning about task-relevant changes relative to the present: what should change versus remain invariant under the instruction.
Without explicitly constructing a current-world knowledge prior $\mathit{W}_t$, the prediction lacks a present-time causal anchor, making relative change reasoning ill-posed and prone to prior-free imagination.
Moreover, even if one predicts differences implicitly, unconstrained continuous variations can be highly scene- and instruction-dependent, and thus are ill-suited as a compact and consistent conditioning interface for policy learning.

Motivated by these issues, we reformulate predictive world modeling as \textbf{prior-grounded variation modeling}.
We first extract the current world knowledge prior $\mathit{W}_t$, and then predict the world-knowledge variation from $t$ to $t{+}n$, denoted as $\Delta\mathit{W}_{t\rightarrow t+n}$.
To provide a policy-usable interface, we represent $\Delta\mathit{W}_{t\rightarrow t+n}$ in a discrete latent form.
Finally, the model generates the action sequence $\mathit{A}_{t:t+n-1}$ conditioned on this discrete variation representation.
\begin{figure*}
        \centering
        \vspace{-5pt}
        \includegraphics[width=1\linewidth]{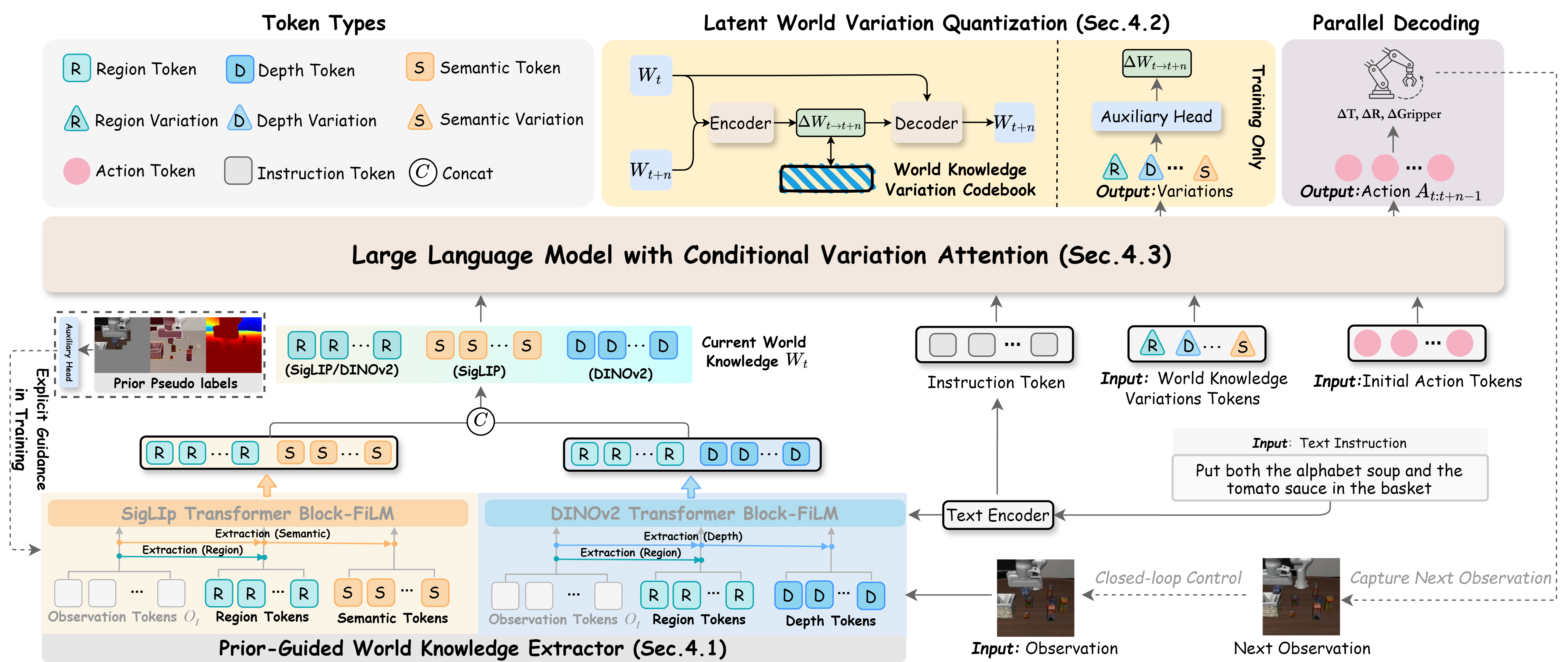}
            \vspace{-15pt}
            \caption{\textbf{Overview of \boldmath$\Delta$VLA Framework.} 
            It first extracts current world knowledge via the \textbf{Prior-Guided World Knowledge Extractor}, which encodes semantic, regional, and depth cues from SigLIP and DINOv2  guided by auxiliary heads and prior pseudo labels to form a unified world prior. 
            These representations are concatenated with world variation tokens and modeled through the \textbf{Conditional Variation Attention} for structured perception–variation reasoning. 
            Finally, the \textbf{Latent World Variation Quantization} module learns a discrete codebook representing world knowledge variations, providing causal guidance for variation token learning and consistent action generation.}
            
    \label{fig:overall_method}
    \vspace{-10pt}
\end{figure*}

\section{\boldmath$\Delta$VLA}
 The framework is shown in Fig.~\ref{fig:overall_method}. We first describe the process of extracting the current world knowledge in Sec.~\ref{World Knowledge Extractor}. 
We then elaborate on how to represent the world variation in Sec.~\ref{World Variation}. Finally, in Sec.~\ref{Conditional Variation Attention}, we introduce the Conditional Variation Attention used during variation modeling.
\subsection{Prior-Guided World Knowledge Extractor}
\label{World Knowledge Extractor}
\begin{figure}
    \centering
    \includegraphics[width=1.0\linewidth]{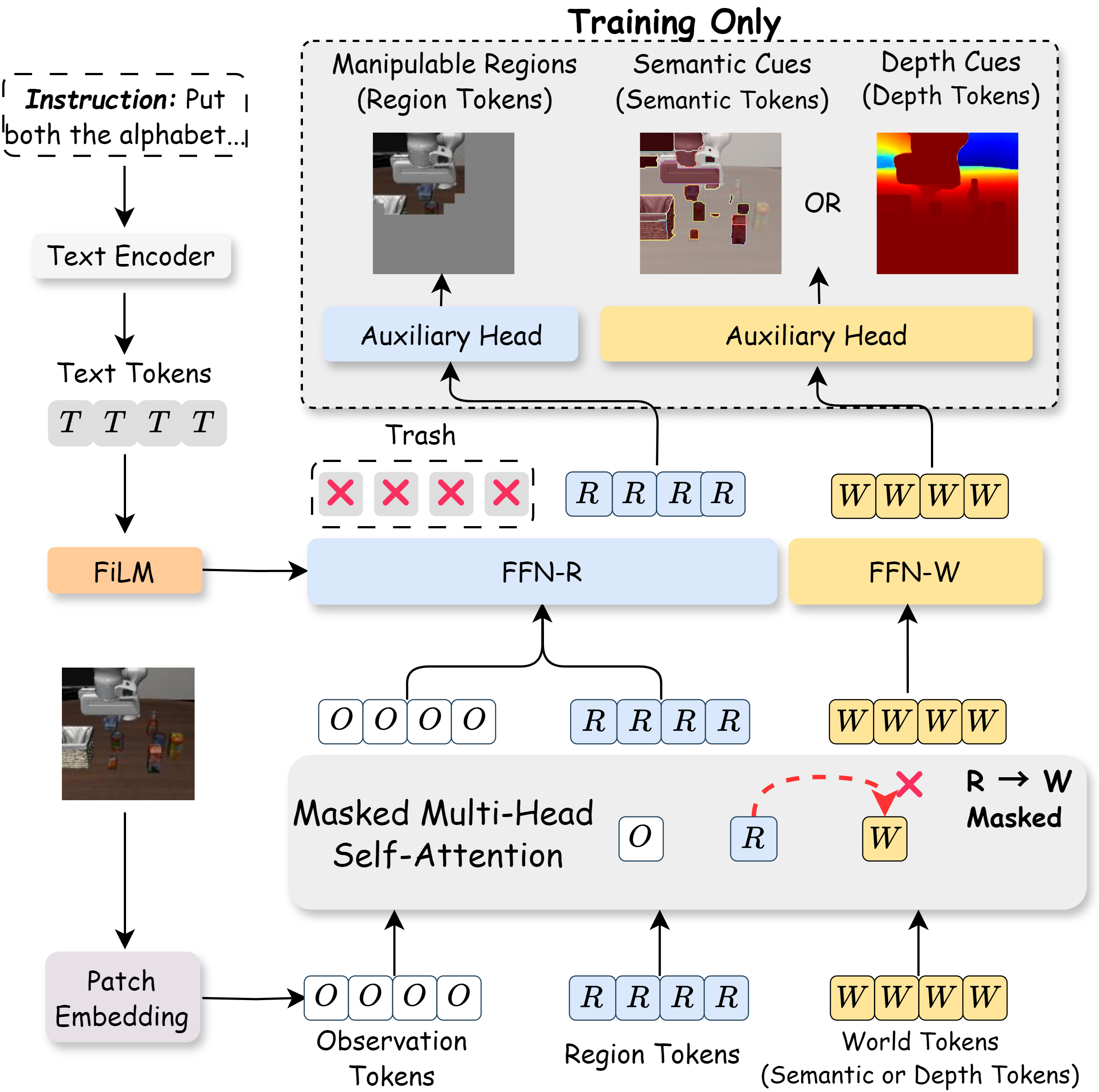}
    \vspace{-15pt}
    \caption{\textbf{Illustration of PWKE.} 
World tokens have different meanings across encoders: they correspond to \textit{semantic tokens} in SigLIP and \textit{depth tokens} in DINOv2. 
The dashed box indicates auxiliary heads used only for training supervision.}

    \label{fig:PWKE}
    \vspace{-12pt}
\end{figure}

\vspace{-1pt}
\noindent\textbf{The Overall Process of PWKE.} 
As shown in Fig.~\ref{fig:overall_method}, 
the PWKE module first constructs the current world knowledge $\mathit{W}_t$ 
from the input $\mathbf{X} = \{\mathit{O}_t, \mathit{I}\}$, 
where $\mathit{O}_t$ is the current visual observation and $\mathit{I}$ is the instruction. 
It leverages the complementary training paradigms of SigLIP (contrastive vision-language alignment) 
and DINOv2 (self-supervised geometry learning) to extract semantic and depth features, 
and aggregates the most likely manipulable regions under the guidance of $\mathit{I}$. 
Specifically, as shown in Fig.~\ref{fig:PWKE}, we introduce two sets of learnable tokens: 
\textit{Region Tokens} $\mathbf{T}_r$ and \textit{World Tokens} $\mathbf{T}_w$, 
to extract different components of the current world knowledge. 
$\mathbf{T}_r$ is designed to localize the most likely manipulable regions, 
while $\mathbf{T}_w$  is used to extract semantic cues from SigLIP and depth cues from DINOv2. World tokens carry encoder-specific meanings: semantic tokens in SigLIP and depth tokens in DINOv2.
We concatenate the observation tokens ${O}_t$, region tokens $\mathbf{T}_r$, and world tokens $\mathbf{T}_w$, 
and feed the combined sequence into the visual encoder blocks for joint representation learning. To keep region tokens focused on original visual observations, 
we mask their attention to world tokens in the self-attention layer:
\begin{equation}
\small
[\tilde{\mathit{O}}_t,\, \tilde{\mathbf{T}}_r,\, \tilde{\mathbf{T}}_w]
= \text{SelfAttn}\big([\mathit{O}_t, \mathbf{T}_r,\mathbf{T}_w],\, \mathbf{M}\big),
\end{equation}
where $[\,\cdot\,, \cdot\,]$ denotes concatenation along sequence dimension and the attention mask $\mathbf{M}$ is defined as:
\begin{equation}
\small
\mathbf{M}_{ij} =
\begin{cases}
-\infty, & i \in \mathcal{I}_r,\, j \in \mathcal{I}_w,\\
0, & \text{otherwise},
\end{cases}
\end{equation}
where $\mathcal{I}_r$ and $\mathcal{I}_w$ denote the index sets of region and world tokens, respectively.

Next, we design two expert FFN branches, 
one for the region-token stream and the other for the world-token stream. 
For the region branch, we further introduce the FiLM modulation mechanism 
that aggregates task-specific manipulable regions guided by the instruction  $\mathit{I}$:
\begin{equation}
\small
\begin{aligned}
f_{\text{R}}\!\big(\mathit{I}, [\tilde{\mathit{O}}_t, \tilde{\mathbf{T}}_r]\big)
&= \big(1 + \gamma(\mathit{I})\big) \odot [\tilde{\mathit{O}}_t, \tilde{\mathbf{T}}_r]
+ \beta(\mathit{I}),\\
[\tilde{\mathit{O}}_t', \tilde{\mathbf{T}}_r']
&= \text{FFN-R}\!\big(f_{\text{R}}(\mathit{I}, [\tilde{\mathit{O}}_t, \tilde{\mathbf{T}}_r])\big),
\end{aligned}
\end{equation}
where $\gamma(\mathit{I})$ and $\beta(\mathit{I})$ are FiLM-generated 
instruction-dependent scale and shift vectors, 
$\odot$ denotes element-wise multiplication, 
and $\text{FFN-R}$ is a region-specific feed-forward expert.
Through iterative visual encoder blocks, region tokens adaptively integrate instruction-relevant information from observation tokens while filtering out redundant details.
Thus, only the final region tokens $\tilde{\mathbf{T}}_r'$ are retained for subsequent processing, and the observation tokens $\tilde{\mathit{O}}_t'$ are discarded. Similarly, for the world-token branch, we apply a dedicated feed-forward expert $\text{FFN-W}$ 
to the self-attention output $\tilde{\mathbf{T}}_w$, 
 whose output $\tilde{\mathbf{T}}_w'$ 
is used for subsequent variation modeling. To explicitly guide the tokens to aggregate their designated knowledge, we introduce token-specific supervision. The processes are described below.

\noindent\textbf{Explicit Supervision for World Knowledge.} 
To obtain manipulable region labels, we derive coarse binary masks from inter-frame motion cues. 
Given consecutive RGB frames of resolution $H \times W$, keypoints are uniformly sampled every 8 pixels, 
yielding $N = \lfloor H/8 \rfloor \times \lfloor W/8 \rfloor$ locations per frame. 
Inter-frame displacements for each sampled point are estimated using CoTracker~\cite{karaev2024cotracker}, 
and a speed threshold $\tau$ is applied to generate a binary motion mask, 
whose corresponding image regions serve as manipulable region labels during training.
For depth and semantic supervision, we employ Depth-Anything v2~\cite{yang2024depth} and SAM~\cite{kirillov2023segment} to generate pseudo ground-truth labels, respectively.

The obtained labels are employed to explicitly guide the learning of  $\tilde{\mathbf{T}}_r'$ and $\tilde{\mathbf{T}}_w'$, 
ensuring that each token aligns with its corresponding world-knowledge type. Specifically, as shown in Fig.~\ref{fig:PWKE}, we use a lightweight Transformer decoder $f_{\text{Dec}}$  that takes $\tilde{\mathbf{T}}_r'$ or $\tilde{\mathbf{T}}_w'$ with a set of learnable mask tokens $\mathbf{T}_{\text{mask}}$ as input, 
and reconstructs corresponding modalities:
\begin{equation}
\small
\hat{\mathbf{Y}} = f_{\text{Dec}}\!\big([\tilde{\mathbf{T}}',\, \mathbf{T}_{\text{mask}}]\big),
\;\text{where}\;
\tilde{\mathbf{T}}' \in \{\tilde{\mathbf{T}}_r',\, \tilde{\mathbf{T}}_w'\}.
\end{equation}
$\hat{\mathbf{Y}}$ denotes the reconstructed modality supervised by the corresponding pseudo ground-truth. We optimize this process using a mean squared error (MSE) loss 
and define it as the current-world knowledge loss $\mathcal{L}_{\text{cur}}$. These decoders serve as auxiliary heads used only during training and do not introduce any additional inference overhead. 
\subsection{Latent World Variation Quantization}
\label{World Variation}
After the PWKE process, we obtain $\tilde{\mathbf{T}}_r'$ and $\tilde{\mathbf{T}}_w'$, 
where $\tilde{\mathbf{T}}_w' \in \{\tilde{\mathbf{T}}_s', \tilde{\mathbf{T}}_d'\}$, 
with $\tilde{\mathbf{T}}_s'$ and $\tilde{\mathbf{T}}_d'$ denoting the semantic tokens from SigLIP 
and the depth tokens from DINOv2, respectively. Subsequently, these tokens are concatenated along the sequence dimension 
to form the current world knowledge representation $\mathit{W}_t$:
\begin{equation}
\small
\mathit{W}_t = [\,\tilde{\mathbf{T}}_r',\, \tilde{\mathbf{T}}_w'\,]
= [\,\tilde{\mathbf{T}}_r',\, \tilde{\mathbf{T}}_s',\, \tilde{\mathbf{T}}_d'\,].
\end{equation}
Building upon this, rather than directly predicting the future world knowledge $\mathit{W}_{t+n}$, 
we model its variation $\Delta\mathit{W}_{t\rightarrow t+n}$. 

Inspired by Genie~\cite{bruce2024genie}, 
we propose the Latent World Variation Quantization (LWVQ) module 
to encode world-knowledge variations in a fully unsupervised manner. 
As illustrated in Fig.~\ref{fig:overall_method}, 
LWVQ follows an encoder–decoder architecture that models the variation of world knowledge 
within a fixed temporal window $n$, corresponding to the action-chunk length. 
During training, the encoder receives the full-modality world-knowledge labels of the current and future frames, $\mathit{W}_t$ and $\mathit{W}_{t+n}$, and encodes their difference into a continuous latent variation.
This continuous representation is subsequently discretized through vector quantization using a learnable 
World Knowledge Variation Codebook. 
The decoder reconstructs the future world knowledge $\mathit{W}_{t+n}$ 
based on the current state $\mathit{W}_t$ and the quantized latent variation $\Delta\mathit{W}_{t\rightarrow t+n}$, 
enabling the model to learn structured world dynamics without explicit future supervision:
\begin{equation}
\small
{\mathit{W}}_{t+n} = 
\text{Dec}\!\Big(
\mathit{W}_t,\,
\underbrace{\text{Quant}\big(\text{Enc}(\mathit{W}_t,\, \mathit{W}_{t+n})\big)}_{\Delta\mathit{W}_{t\rightarrow t+n}}
\Big),
\end{equation}
where $\text{Enc}(\cdot)$ and $\text{Dec}(\cdot)$ denote the encoder and decoder, respectively, 
and $\text{Quant}(\cdot)$ represents vector quantization parameterized by the codebook.

\noindent\textbf{The Optimization Process of LWVQ.} The LWVQ is optimized following the VQ-VAE objective~\cite{van2017neural}. 
During training, the encoder outputs continuous embeddings that are discretized by mapping them 
to the closest entries within a learnable codebook, where each vector represents a distinct prototype of world-knowledge variation. 
After training, the encoder is employed to compute the variation of each knowledge type, 
which serves as the supervision for subsequent variation modeling in $\Delta$VLA.
This formulation transforms continuous features into discrete latent, 
providing a compact and structured representation that facilitates downstream prediction. Compared to full future modalities prediction, this design yields a compact and interpretable representation of world dynamics, 
facilitating efficient downstream reasoning and robust generalization.
\subsection{Conditional Variation Attention}
\label{Conditional Variation Attention}
\begin{figure}
    \centering
    \includegraphics[width=1\linewidth]{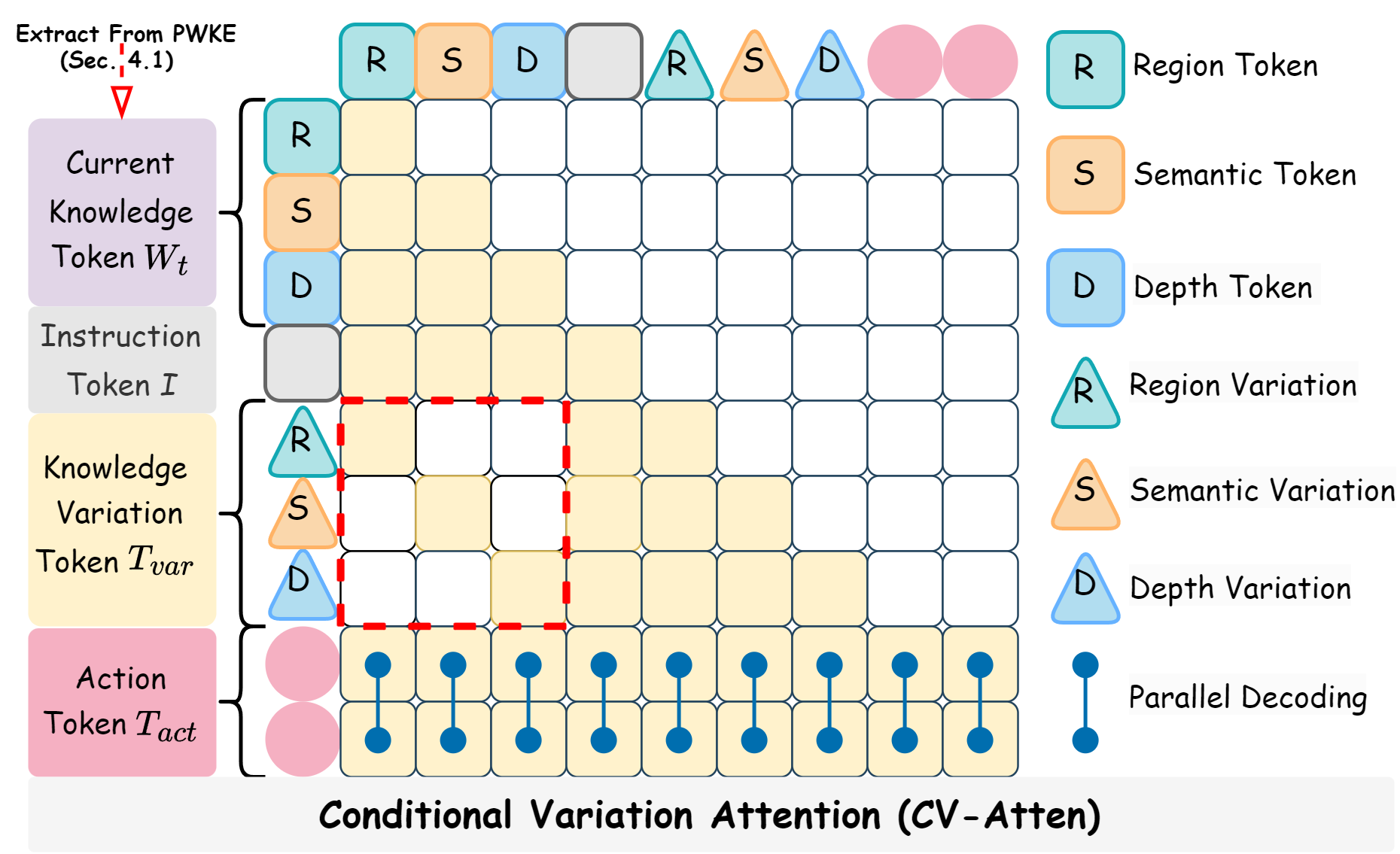}
    \vspace{-15pt}
    \caption{\textbf{Illustration of CV-Atten.} 
In red dashed boxes, each variation token attends to its corresponding world prior, ensuring type-specific modeling and minimizing cross-modality leakage. For action generation, we employ parallel decoding.}
    \label{fig:mask}
    \vspace{-13pt}
\end{figure}
After encoding the variation of each world-knowledge modality through the LWVQ module, 
we introduce a learnable set of world-variation tokens to reason over these variations within a unified latent space. 
Then, the current world knowledge $\mathit{W}_t$, the instruction $\mathit{I}$, 
the variation tokens $\mathbf{T}_{\textit{var}}$, and a set of initialized action tokens $\mathbf{T}_{\textit{act}}$ 
are concatenated and fed into the large language model $f_{\text{LLM}}$ for unified reasoning. 
The process is formulated as:
\begin{equation}
\small
[\tilde{\mathbf{T}}_{\textit{var}}',\, \mathit{A}_{t:t+n-1}] =
f_{\text{LLM}}\big([\mathit{W}_t,\, \mathit{I},\, \mathbf{T}_{\textit{var}},\, \mathbf{T}_{\textit{act}}], \mathbf{M}_{cv}\big),
\label{eq:cv-atten}
\end{equation}
where $f_{\text{LLM}}$ denotes the large language model that outputs 
the variation representations $\tilde{\mathbf{T}}_{\textit{var}}'$ and predicts the $n$-step action sequence $\mathit{A}_{t:t+n-1}$. 

Specifically, to reduce cross-modality leakage and enforce type-specific variation modeling, 
we introduce a Conditional Variation Attention (CV-Atten) mechanism, denoted as $\mathbf{M}_{cv}$ in Eq.~\ref{eq:cv-atten}.
As illustrated in Fig.~\ref{fig:mask}, CV-Atten adopts a structured attention masking strategy 
that restricts each variation token to attend exclusively to its corresponding world-knowledge prior 
(e.g., semantic-to-semantic, depth-to-depth, region-to-region). 
This design enforces independent variation reasoning across modalities, 
preventing semantic and geometric interference while maintaining global consistency through shared normalization. 
As demonstrated in our experiments, 
this mechanism significantly improves cross-modal disentanglement and leads to more consistent and interpretable action generation.

To further guide variation learning, we attach an auxiliary supervision head to 
$\tilde{\mathbf{T}}_{\textit{var}}'$ during training. 
The corresponding ground-truth variations $\Delta\mathit{W}_{t\rightarrow t+n}$ are obtained from Sec.~\ref{World Variation}, 
based on which we define the variation reconstruction loss $\mathcal{L}_{\text{var}}$. The optimization process follows a similar approach to $\mathcal{L}_{\text{cur}}$, using MSE loss.
During parallel decoding, the predicted $n$-step action sequence $\mathit{A}_{t:t+n-1}$ is optimized with an L1 loss $\mathcal{L}_{\text{act}}$.
The entire $\Delta$VLA framework is trained with three components: 
the current-world knowledge reconstruction loss $\mathcal{L}_{\text{cur}}$, 
the variation modeling loss $\mathcal{L}_{\text{var}}$, 
and the action prediction loss $\mathcal{L}_{\text{act}}$. 
The complete training procedure of $\Delta$VLA is summarized in Alg.~\ref{alg:deltavla}.

\begin{algorithm}[t]
\small
\caption{Training Pipeline of $\Delta$VLA}
\label{alg:deltavla}
\begin{algorithmic}[1]
\REQUIRE Observation $O_t$, instruction $I$, future observation $O_{t+n}$, action sequence $A_{t:t+n-1}$

\STATE \textcolor{gray}{\textbf{\textit{// Stage 1: Train PWKE for World Knowledge Extraction}}}

\STATE $W_t \leftarrow \text{PWKE}(O_t, I)$
\STATE $\mathcal{L}_{cur} \leftarrow \mathcal{L}_{cur}(W_t)$
\STATE Update PWKE parameters $\theta_{\text{PWKE}}$ via $\nabla \mathcal{L}_{cur}$

\STATE \textcolor{gray}{\textbf{\textit{// Stage 2: Pretrain LWVQ for World Variation Quantization}}}

\STATE $W_t \leftarrow \text{PWKE}(O_t, I)$
\STATE $W_{t+n} \leftarrow \text{PWKE}(O_{t+n}, I)$
\STATE $z \leftarrow \text{Enc}(W_t, W_{t+n})$
\STATE $\Delta W_{t\rightarrow t+n} \leftarrow \text{Quant}(z)$
\STATE $\hat{W}_{t+n} \leftarrow \text{Dec}(W_t, \Delta W_{t\rightarrow t+n})$
\STATE $\mathcal{L}_{\text{LWVQ}} \leftarrow \|W_{t+n}-\hat{W}_{t+n}\|^2$
\STATE Update LWVQ parameters $\theta_{\text{LWVQ}}$ via $\nabla \mathcal{L}_{\text{LWVQ}}$

\STATE \textcolor{gray}{\textbf{\textit{// Stage 3: Train $\Delta$VLA with Frozen PWKE and LWVQ}}}

\STATE $W_t \leftarrow \text{PWKE}(O_t, I)$
\STATE $W_{t+n} \leftarrow \text{PWKE}(O_{t+n}, I)$
\STATE $\Delta W_{t\rightarrow t+n}^{*} \leftarrow \text{Quant}(\text{Enc}(W_t, W_{t+n}))$
\STATE Initialize variation tokens $T_{var}$ and action tokens $T_{act}$
\STATE $[\tilde T_{var}, A_{t:t+n-1}] \leftarrow f_{LLM}([W_t, I, T_{var}, T_{act}], M_{cv})$
\STATE $\mathcal{L}_{var} \leftarrow \|\tilde T_{var} - \Delta W_{t\rightarrow t+n}^{*}\|^2$
\STATE $\mathcal{L}_{act} \leftarrow \|A_{t:t+n-1} - A_{t:t+n-1}^{*}\|_1$
\STATE $\mathcal{L} \leftarrow \mathcal{L}_{act} + \lambda_{var}\mathcal{L}_{var}$
\STATE Update $\Delta$VLA parameters via $\nabla \mathcal{L}$

\end{algorithmic}
\end{algorithm}
\section{Experiments}
\subsection{Experimental Setup}
\begin{table*}[t]
    \caption{\textbf{Simulation Experimental Results on the LIBERO Benchmark.} Comparison of task success rates (SR) and their ranks (RK). ``\dag'' indicates our reproduced results. Bold indicates the highest score.}
    \label{tab:libero-p}
    \centering
    \footnotesize
    \setlength{\tabcolsep}{10pt}
    \begin{tabular}{l|cc|cc|cc|cc|cc}
    \toprule
    \textbf{Method} & \multicolumn{2}{c|}{\textbf{Spatial}} & \multicolumn{2}{c|}{\textbf{Object}} & \multicolumn{2}{c|}{\textbf{Goal}} & \multicolumn{2}{c|}{\textbf{Long}} & \multicolumn{2}{c}{\textbf{Average}} \\ 
    & SR $\uparrow$ & RK $\downarrow$ & SR $\uparrow$ & RK $\downarrow$ & SR $\uparrow$ & RK $\downarrow$ & SR $\uparrow$ & RK $\downarrow$ & SR $\uparrow$ & RK $\downarrow$ \\
    \midrule
    OpenVLA~\textit{[CoRL'24]}~\cite{openvla} & 84.7 & 10 & 88.4 & 10 & 79.2 & 10 & 53.7 & 12 & 76.5 & 10 \\
    MDT~\textit{[RSS'24]}~\cite{MDT} & 78.5 & 11 & 87.5 & 11 & 73.5 & 11 & 64.8 & 11 & 76.1 & 11 \\
     $\pi_0$~\textit{[RSS'25]}~\cite{pi0} & 96.8 & 5 & 98.8 & 2 & 95.8 & 3 & 85.2 & 9 & 94.2 & 7 \\
    OpenVLA-OFT~\textit{[RSS'25]}~\cite{oft} & 97.6 & 3 & 98.4 & 3 & \textbf{97.9} & \textbf{1} & 94.5 & 2 & 97.1 & 2 \\
    UniVLA~\textit{[RSS'25]}~\cite{univla} & 96.5 & 6 & 96.8 & 6 & 95.6 & 4 & 92.0 & 3 & 95.2 & 4 \\
    PD-VLA~\textit{[arXiv'25]}~\cite{pdvla} & 95.5 & 7 & 96.7 & 7 & 94.9 & 7 & 91.7 & 4 & 94.7 & 5 \\
    CoT-VLA~\textit{[CVPR'25]}~\cite{cot} & 87.5 & 9 & 91.6 & 9 & 87.6 & 9 & 69.0 & 10 & 83.9 & 9 \\
    Seer~\textit{[ICLR'25]}~\cite{seer} & - & - & - & - & - & - & 87.7 & 8 & - & - \\
    STAR~\textit{[ICML'25]}~\cite{star} & 95.5 & 7 & 98.3 & 4 & 95.0 & 6 & 88.5 & 7 & 94.3 & 6 \\
    DreamVLA~\textit{[NeurIPS'25]}~\cite{zhang2025dreamvla} & 97.5 & 4 & 94.0 & 8 & 89.5 & 8 & 89.5 & 6 & 92.6 & 8 \\
    $\mathcal{F}_1$~\textit{[arXiv'25]}~\cite{lv2025f1} & 98.2 & 2 & 97.8 & 5 & 95.4 & 5 & 91.3 & 5 & 95.7 & 3 \\
    \rowcolor[HTML]{FFF5CC}  
    \boldmath$\Delta$VLA & \textbf{98.6} & \textbf{1} & \textbf{99.4} & \textbf{1} & 97.4 & 2 & \textbf{95.6} & \textbf{1} & \textbf{97.8} & \textbf{1} \\
    \bottomrule
    \end{tabular}
    \vspace{-5pt}
\end{table*}
\begin{table*}[t]
\caption{\textbf{Simulation Experimental Results on the RoboTwin~2.0 Benchmark.} 
Comparison of success rates across eight tasks, along with the overall average performance.
``\dag'' indicates our reproduced results. Bold indicates the highest score.}
\label{tab:robotwin-comparison}
\centering
\footnotesize
\setlength{\tabcolsep}{6pt}
\begin{tabular}{l|cccccccc|c}
\toprule
\textbf{Method} & \textbf{Click A.} & \textbf{Open L.} & \textbf{Place C. P.} & \textbf{Dump B. B.} & \textbf{Press S.} & \textbf{Click B.} & \textbf{Stack B. T.} & \textbf{Place B. S.} & \textbf{Average} \\
\midrule
RDT~\cite{liu2024rdt} & 61.0 & 59.0 & 78.0 & 64.0 &41.0 &80.0&76.0 &5.0  & 58.0 \\
ACT~\cite{zhao2023learning} & 32.0 & 56.0 & 72.0 &68.0 &31.0  &58.0 &82.0  &7.0  &50.8  \\
Diffusion Policy~\cite{chi2025diffusion} & 61.0 & 49.0 & 41.0 &49.0 &6.0 &54.0  &61.0 &11.0  &41.5   \\
3D Diffusion Policy~\cite{ze20243d} & 77.0 & 82.0 & 86.0 &85.0  &69.0 &90.0 &83.0  &19.0 &73.9 \\
 $\pi_0$~\cite{pi0} & 63.0 & 85.0 & 88.0 &83.0  &62.0  &44.0 &\textbf{91.0} &23.0 &67.4 \\
OpenVLA-OFT\dag~\cite{oft} & 82.0 &73.0  &89.0  &86.0  &68.0  &84.0  &83.0  &13.0  &72.3  \\
 \rowcolor[HTML]{FFF5CC}  
\boldmath$\Delta$VLA & \textbf{91.0} & \textbf{86.0} & \textbf{92.0} & \textbf{91.0} & \textbf{78.0} &\textbf{93.0}  &87.0  &\textbf{25.0}  &\textbf{80.4}  \\
\bottomrule
\end{tabular}
\vspace{-10pt}
\end{table*}

\noindent\textbf{Implementation Details.} 
We train on 8×A800~(80GB) GPUs and perform real-world inference on a single RTX 5090 (32GB) GPU. For simulation, we build on OpenVLA~\cite{kim2024openvla} as the backbone. For the LIBERO~\cite{libero}, the network is fine-tuned using Low-Rank Adaptation (LoRA) with rank 32 and scaling factor $\alpha=64$ for 60K optimization steps, with a batch size of 64 and an initial learning rate of $5\times10^{-4}$. We set the action chunk length to $K=8$. We evaluate checkpoints every 10K steps on the validation benchmarks and report results using the checkpoint that achieves the best performance. For the RoboTwin 2.0~\cite{chen2025robotwin}, we set a batch size of 32, an initial learning rate of $5 \times 10^{-4}$, and an action chunk size of 25. As for the Region and World tokens, we set their numbers to 64 and 9, respectively. 
For real-world experiments, we adopt an action chunk length of $K=25$ and fine-tune OpenVLA using LoRA with rank 32 and $\alpha=64$. The model is trained for 80K steps with a batch size of 32, starting from a learning rate of $5\times10^{-4}$, which is reduced to $5\times10^{-5}$ after 50K steps. Beginning at step 60K, we evaluate checkpoints every 10K steps and use the best-performing checkpoint. 

For selecting manipulable regions, we apply a speed threshold $\tau$ to generate a binary motion mask, and set $\tau = 1$. For depth supervision, we use monocular depth estimators, specifically Depth-Anything v2~\cite{yang2024depth}, to generate pseudo-ground-truth depth labels. For semantic information, we adopt SAM~\cite{kirillov2023segment} and use the output of its image encoder as dense, segmentation-aware features. 
For the LWVQ, we utilize the C-ViViT model architecture from Villegas~\cite{villegas2022phenaki} to replicate the model from GENIE~\cite{bruce2024genie}. 
We use a quantization approach for the world knowledge variation codebook with a codebook size of 8 and a quantization dimension of 32.
During training, we use a batch size of 64 and set the learning rate to $1\times10^{-4}$ for a total of 30K steps. 

\noindent\textbf{Simulation Benchmark.} 
We conduct extensive evaluations on two simulation benchmarks:  
(1) Four task suites of \textbf{LIBERO~\cite{libero}—Spatial, Object, Goal, and Long}, each with 10 tasks and 50 demonstrations; and  
(2) Eight task configurations from the \textbf{RoboTwin~2.0~\cite{chen2025robotwin}}.  LIBERO features substantially richer natural-language instructions than RLBench~\cite{james2020rlbench}, making it a challenging testbed for language grounding and multimodal reasoning. For LIBERO, we evaluate each task over 50 trials.
RoboTwin 2.0~\cite{chen2025robotwin} provides a large-scale and highly diverse benchmark for evaluating bimanual robotic manipulation under realistic and domain-randomized settings. We select eight tasks for evaluation: 1) Click Alarmclock, 2) Open Laptop, 3) Place Container Plate, 4) Dump Bin Bigbin, 5) Press Stapler, 6) Click Bell, 7) Stack Bowls Two, and 8) Place Bread Skillet. Each task is evaluated over 100 trials.

\noindent\textbf{Real-World Experiments.} We deploy $\Delta$VLA on both the AgileX Cobot Magic~\cite{aloha} and Galaxea R1 Lite platforms~\cite{galaxea} for real-world evaluations. We conduct four long-horizon tasks: 1) Drawer Manipulation, 2) Align Shoes, 3) T-shirt Folding, and 4) Plate Lemon Apple. Each task contains 50 human-teleoperated demonstrations.

\noindent\textbf{Baselines.} We compare $\Delta$VLA with a range of state-of-the-art methods, including general VLA models such as OpenVLA~\cite{openvla}, OpenVLA-OFT~\cite{oft}, STAR~\cite{star}, $\pi_0$~\cite{pi0}, and PD-VLA~\cite{pdvla}. 
In addition, we evaluate against predictive-paradigm approaches such as CoT-VLA~\cite{cot}, Seer~\cite{seer}, MDT~\cite{MDT}, UniVLA~\cite{univla}, $\mathcal{F}_1$~\cite{lv2025f1}, and DreamVLA~\cite{zhang2025dreamvla}, 
which explicitly incorporate future subgoal imagery or world-knowledge foresight into action reasoning. For real-world experiments, we reproduce representative models including OpenVLA~\cite{openvla}, OpenVLA-OFT~\cite{oft}, Seer~\cite{seer}, and DreamVLA~\cite{zhang2025dreamvla}, 
using the same deployment settings.
\vspace{-5pt}
\subsection{Overall Performance}
\vspace{-2pt}
\noindent\textbf{Simulation Experimental Results.}
We evaluate $\boldsymbol{\Delta}$VLA on two challenging simulation benchmarks, LIBERO and RoboTwin 2.0.
As reported in Tables~\ref{tab:libero-p} and~\ref{tab:robotwin-comparison},
$\boldsymbol{\Delta}$VLA achieves a success rate of 97.8\% on LIBERO and 80.4\% on RoboTwin 2.0, establishing new state-of-the-art performance across diverse task suites.
Compared to predictive paradigms such as CoT-VLA~\cite{cot} and DreamVLA~\cite{zhang2025dreamvla}, which learn to forecast future states and then derive actions from the predicted outcomes, $\boldsymbol{\Delta}$VLA yields consistently higher success rates across both benchmarks.
This suggests that simply generating plausible future observations is not sufficient for robust control, especially when the predicted futures are weakly grounded in the current interaction context.
In contrast, our method explicitly models world-knowledge variations that capture action-induced, decision-relevant changes, which provides a more direct learning signal for manipulation policies.

Moreover, $\boldsymbol{\Delta}$VLA also outperforms non-predictive VLA baselines such as OFT~\cite{oft}, indicating that the gains are not merely attributed to introducing an auxiliary prediction objective.
Instead, the improvements stem from reformulating prediction as learning task-relevant variations and aligning them with action generation.
Overall, $\boldsymbol{\Delta}$VLA bridges the gap between perception and control more effectively than prior approaches, leading to stronger robustness under environment variations and better generalization.
\begin{table*}[t]
\caption{\textbf{Real-World  Results.} ``\dag'' indicates results reproduced using the same settings as \boldmath$\Delta$VLA. The table shows success rates for the progressive stage across four long-horizon tasks, with the average success rate representing only the completion of the entire task.}
    \label{tab:real_world}
    \centering
    \footnotesize
    \setlength{\tabcolsep}{3.5pt}
    \begin{tabular}{l|ccc|ccc|ccc|ccc|c}
    \toprule
    \textbf{Method} & \multicolumn{3}{c|}{\textbf{Drawer Manipulation}} & \multicolumn{3}{c|}{\textbf{Align Shoes}} & \multicolumn{3}{c|}{\textbf{T-shirt Folding}}  & \multicolumn{3}{c|}{\textbf{Plate Lemon Apple}} & \textbf{Average}\\ 
     & \textbf{Pull} & \textbf{+Place} & \textbf{+Push} & \textbf{Retrieve}  & \textbf{+Retrieve} & \textbf{+Align} & \textbf{Step 1} & \textbf{+Step 2} & \textbf{+Step 3}  & \textbf{Open} & \textbf{+Retrieve} & \textbf{+Cover} & \\
    \hline
    \rowcolor[HTML]{EFEFEF} \multicolumn{14}{c}{\textit{\scriptsize Galaxea R1 Lite platform}} \\
    \hline
    OpenVLA\dag~\cite{openvla} &15/25 &13/25 &10/25 &13/25&12/25&10/25&10/25&8/25&6/25&14/25&13/25&10/25&36\% \\
    OpenVLA-OFT\dag~\cite{oft} &18/25 &17/25&16/25 &16/25&16/25&15/25&12/25&10/25&8/25&13/25&12/25&12/25&51\% \\
    UniVLA\dag~\cite{univla} &16/25 &14/25&13/25 &15/25&14/25&14/25&15/25&14/25&12/25&11/25&10/25&9/25&48\% \\
     Seer\dag~\cite{seer} &14/25 &14/25&12/25 &16/25&13/25&10/25&11/25&11/25&11/25&15/25&13/25&12/25&45\% \\
     DreamVLA\dag~~\cite{zhang2025dreamvla} &15/25 &15/25&14/25 &17/25&14/25&14/25&13/25&11/25&11/25&15/25&14/25&14/25&53\% \\
    
        \rowcolor[HTML]{FFF5CC}  
    \boldmath$\Delta$VLA  &\textbf{23/25} & \textbf{22/25}&\textbf{20/25} &\textbf{20/25}&\textbf{19/25}&\textbf{19/25}&\textbf{17/25}&\textbf{17/25}&\textbf{16/25}&\textbf{19/25}&\textbf{19/25}&\textbf{17/25}&\textbf{72\%} \\
    \hline
    \rowcolor[HTML]{EFEFEF} 
    \multicolumn{14}{c}{\textit{\scriptsize AgileX Cobot Magic platform}} \\
    \hline
    OpenVLA-OFT\dag~\cite{oft} &18/25 &17/25&16/25 &17/25&16/25&15/25&12/25&11/25&11/25&14/25&12/25&10/25&52\% \\
     DreamVLA\dag~~\cite{zhang2025dreamvla} &15/25 &14/25&13/25 &16/25&15/25&13/25&14/25&11/25&11/25&14/25&13/25&12/25&49\% \\
    \rowcolor[HTML]{FFF5CC}  
    \boldmath$\Delta$VLA  &\textbf{23/25} &\textbf{21/25} &\textbf{18/25} &\textbf{19/25}&\textbf{19/25}&\textbf{18/25}&\textbf{19/25}&\textbf{17/25}&\textbf{15/25}&\textbf{20/25}&\textbf{19/25}&\textbf{18/25}&69\% \\
    \bottomrule
    \end{tabular}
    \vspace{-8pt}
\end{table*}

\begin{table}[t]
\caption{\textbf{Efficiency Comparison Results.} \textit{Latency} denotes inference delay; \textit{Throughput} denotes the number of predicted actions per second; \textit{Cost} is training time per 10k steps. ``\dag''  denotes results reproduced under identical settings as $\Delta$VLA. All results are reported on A800 GPUs for a fair comparison.}
    \label{tab:efficiency}
    \centering
    \footnotesize
    \setlength{\tabcolsep}{1pt}  
    \begin{tabular}{l|cccc}
    \toprule
    \textbf{Method} & \textbf{Latency} $\downarrow$ & \textbf{Throughput} $\uparrow$  & \textbf{Cost} $\downarrow$ & \textbf{LIBERO SR} $\uparrow$ \\
    \midrule
    OpenVLA\dag~\cite{openvla}  & 0.254 s & 3.9 Hz  & 11.7 h & 76.5\% \\
    OpenVLA-OFT\dag~\cite{oft}& 0.132 s & 60.6 Hz & 12.5 h & 97.1\%\\
    PD-VLA\dag~\cite{pdvla} & 0.143 s & 55.9 Hz & 11.7 h  & 94.7\% \\
      Seer\dag~\cite{seer} &0.125 s  &8.0 Hz  &14.3 h & -\\
        DreamVLA\dag~\cite{zhang2025dreamvla} &0.130 s  &7.7 Hz  &14.5 h  & 92.6\% \\
    \midrule
    \rowcolor[HTML]{FFF5CC}  
    \boldmath$\Delta$VLA &\textbf{0.105 s} &\textbf{76.2 Hz}  &\textbf{4.9 h} & \textbf{97.8\%} \\
    \bottomrule
    \end{tabular}
\vspace{-5pt}
\end{table}

\noindent\textbf{Real-world Experimental Results.} 
We evaluate $\boldsymbol{\Delta}$VLA on four challenging long-horizon tasks using the Galaxea R1 Lite and AgileX Cobot Magic platforms. 
As shown in Table~\ref{tab:real_world}, $\boldsymbol{\Delta}$VLA outperforms other state-of-the-art models across all tasks, demonstrating consistent advantages in real-world long-sequence manipulation.
In particular, tasks such as Drawer Manipulation and T-shirt Folding require multi-step decision making with tight geometric constraints and frequent contact state changes, where errors can easily accumulate over time.
$\boldsymbol{\Delta}$VLA achieves strong performance on these tasks, suggesting that modeling action-induced world changes provides decision-relevant signals that help maintain coherent progress across extended horizons, rather than relying on appearance-level predictions alone.
Across platforms, $\boldsymbol{\Delta}$VLA attains an average success rate of 72\% on Galaxea R1 Lite and 69\% on AgileX Cobot Magic, indicating robust transfer under different embodiments, sensing conditions, and execution noise.
Overall, these results highlight the practicality of $\boldsymbol{\Delta}$VLA for real-world deployment, where reliable reasoning about world changes and long-horizon control are both critical for successful task execution.

\noindent\textbf{Efficiency Comparison Results.} 
As shown in Table~\ref{tab:efficiency}, $\boldsymbol{\Delta}$VLA achieves superior efficiency in terms of latency, throughput, and training cost while preserving strong task performance.
It reaches a latency of 0.105 seconds and a throughput of 76.2 Hz, enabling responsive closed-loop control for manipulation.
Meanwhile, $\boldsymbol{\Delta}$VLA maintains a 97.8\% success rate on LIBERO with a training cost of 4.9 hours per 10k steps, indicating a favorable accuracy--efficiency trade-off rather than sacrificing performance for speed.

We attribute these gains to two design choices.
First, PWKE reduces redundant perceptual tokens and retains only decision-critical world knowledge, which lowers the computational burden in both training and inference.
Second, LWVQ learns compact discrete variation codes to represent action-induced world changes, avoiding the expensive requirement of predicting full high-dimensional modalities.
Together, these components make $\boldsymbol{\Delta}$VLA more compute-efficient and easier to scale, particularly in long-horizon settings where latency and throughput directly impact stability and success.
\subsection{Ablation Studies}
\begin{table}[t]
\caption{\textbf{Ablation Study on Model Components.} \textit{LWVQ} denotes the use of latent world variation prediction.}
    \label{tab:Components}
    \centering
    \footnotesize
    \setlength{\tabcolsep}{5pt}  
    \begin{tabular}{ccc|ccccc}
    \toprule
    \textbf{PWKE} & \textbf{LWVQ} & \textbf{CV-Atten} &\textbf{Spatial} & \textbf{Object}  & \textbf{Goal} & \textbf{Long} \\
    \midrule
     & & &95.2  &96.4 &95.4 &91.8 \\
     \checkmark & & &96.0  &97.4 &96.6 &92.6 \\
          &  \checkmark  &  &96.4&97.0&96.2&93.0 \\
     \checkmark & \checkmark &&98.0&98.6&96.6&94.8 \\
         \rowcolor[HTML]{FFF5CC}  
    \checkmark & \checkmark & \checkmark  &\textbf{98.6}&\textbf{99.4}&\textbf{97.4} &\textbf{95.6} \\
    \bottomrule
    \end{tabular}
    \vspace{-5pt}
\end{table}
\begin{table}[t]
\footnotesize
\setlength{\tabcolsep}{2.3pt}
\centering
\caption{\textbf{Ablation Study on PWKE.} \textit{w/ FiLM} indicates the use of textual guidance in extracting manipulable regions.}
\label{tb:ab_PKWE}
\begin{tabular}{cc|cc|cccc}
\toprule
\multicolumn{2}{c|}{\textbf{Region}} &\textbf{Semantic} & \textbf{Depth} &\textbf{Spatial} & \textbf{Object}  & \textbf{Goal} & \textbf{Long} \\
w/ FiLM &w/o FiLM  &  &  &  &  \\
\hline
 \checkmark &  & & &95.6&97.2&96.4&93.4  \\
  \checkmark &  &\checkmark & &96.8&98.0&96.8 &93.8 \\
   \checkmark &  & &\checkmark &97.0&98.2&96.6 &94.4 \\
     \rowcolor[HTML]{FFF5CC}  
    \checkmark &  &\checkmark &\checkmark &\textbf{98.6}&\textbf{99.4}&\textbf{97.4} &\textbf{95.6} \\
    \hline
      & \checkmark  &\checkmark &\checkmark &97.4&98.2&96.2 &94.0 \\
\bottomrule
\end{tabular}
\vspace{-5pt}
\end{table}
\begin{table}[t]
\caption{\textbf{Ablation Study on the Region and World Tokens in PWKE.} We analyze how varying the number of tokens and their attention modeling influences performance.}
\label{tab:Region and World Tokens}
\centering
\footnotesize
\setlength{\tabcolsep}{10.1pt}
\begin{tabular}{c|cccc}
\toprule
\textbf{Method} & \textbf{Spatial} & \textbf{Object} & \textbf{Goal} & \textbf{Long} \\
  \hline
\rowcolor[HTML]{EFEFEF} \multicolumn{5}{c}{\textit{The Number of Region (R) and World (W) Tokens}} \\
\hline
(R:128, ~W:4) & 98.4&98.6&96.6 &95.2 \\
 (R:128, ~W:9) &98.4 &98.8 &97.0 &94.8 \\
(R:64, ~W:4) &98.0 &99.0 &97.2 &94.8\\
   \rowcolor[HTML]{FFF5CC}  
(R:64, ~W:9) &\textbf{98.6}&\textbf{99.4}&\textbf{97.4} &\textbf{95.6}  \\
(R:64, ~W:16)& 98.2&98.8 &96.8 &95.0 \\
\hline
\rowcolor[HTML]{EFEFEF} \multicolumn{5}{c}{\textit{The Attention Modeling}} \\
\hline
Bidirectional &97.6 &98.2 &96.2 &94.4 \\
   \rowcolor[HTML]{FFF5CC}  
Masked R$\rightarrow$W  &\textbf{98.6}&\textbf{99.4}&\textbf{97.4} &\textbf{95.6}  \\
\bottomrule

\end{tabular}
\vspace{-5pt}
\end{table}
\begin{table}[t]
\caption{\textbf{Robustness Evaluation with Noisy Prior Labels.}
We evaluate the effect of injecting different levels of noise into the prior labels.
The \textit{w/o noise} row provides the clean-label.}
\vspace{-5pt}
\label{tab:Noisy Label}
\centering
\footnotesize
\setlength{\tabcolsep}{12pt}
\begin{tabular}{c|cccc}
\toprule
\textbf{Method} & \textbf{Spatial} & \textbf{Object} & \textbf{Goal} & \textbf{Long} \\
\hline
w/o noise &\textbf{98.6}&\textbf{99.4}&\textbf{97.4} &\textbf{95.6}   \\
w/ 10\% &98.4 &99.4 &97.2 &95.4 \\
w/ 20\% &98.0 &99.0 &96.8 &95.4   \\
w/ 30\% &98.0 &98.6 &96.6 &95.0  \\
w/ 50\% &97.0 &97.4 &96.0 &93.8 \\
\bottomrule
\end{tabular}
\vspace{-8pt}
\end{table}
\begin{table}[t]
\caption{\textbf{Ablation Study on the LWVQ Module.}
We investigate the representation of future states and the design of latent variation, comparing shared versus separated settings.}
\label{tab:LWVQ}
\centering
\footnotesize
\setlength{\tabcolsep}{8.1pt}
\begin{tabular}{c|cccc}
\toprule
\textbf{Method} & \textbf{Spatial} & \textbf{Object} & \textbf{Goal} & \textbf{Long} \\
  \hline
\rowcolor[HTML]{EFEFEF} \multicolumn{5}{c}{\textit{Representation of Future States}} \\
\hline
Full Future Modalities &95.6 &96.8 &95.6 &92.0 \\
Continuous Variation &96.8 &97.4 &96.4 &93.2 \\
   \rowcolor[HTML]{FFF5CC}  
Latent Variation &\textbf{98.6} &\textbf{99.4} &\textbf{97.4} &\textbf{95.6} \\
\hline
\rowcolor[HTML]{EFEFEF} \multicolumn{5}{c}{\textit{Shared or Separated Variation Tokens}} \\
\hline
Shared &96.8 &98.6 &96.0 &94.4 \\
   \rowcolor[HTML]{FFF5CC}  
Separated &\textbf{98.6} &\textbf{99.4} &\textbf{97.4} &\textbf{95.6} \\
\bottomrule

\end{tabular}
\vspace{-6pt}
\end{table}

We conduct ablation studies on the LIBERO benchmark to validate the effectiveness of our method.

\noindent\textbf{Ablation Study on Model Components.}
Table~\ref{tab:Components} summarizes the contribution of each component on the four LIBERO suites.
The baseline variant without PWKE, LWVQ, and CV-Atten attains 95.2 on Spatial, 96.4 on Object, 95.4 on Goal, and 91.8 on Long, and the noticeably lower performance on Long confirms that long-horizon manipulation is the most challenging setting.
Introducing PWKE yields consistent improvements across suites, resulting in 96.0 on Spatial, 97.4 on Object, 96.6 on Goal, and 92.6 on Long, which indicates that explicit prior extraction enhances perceptual grounding by filtering redundant observations.
Enabling LWVQ alone also improves performance to 96.4 on Spatial, 97.0 on Object, 96.2 on Goal, and 93.0 on Long, suggesting that learning compact representations of action-induced variations provides an effective supervisory signal, particularly for sustained multi-step execution.
Crucially, PWKE and LWVQ exhibit strong complementarity and deliver the largest gains when jointly applied.
This observation supports our design rationale that PWKE supplies decision-critical interaction priors, while LWVQ captures how such priors evolve under actions, and the two together form a more complete and controllable representation for policy learning.
Finally, incorporating CV-Atten on top of PWKE and LWVQ further improves the results to 98.6 on Spatial, 99.4 on Object, 97.4 on Goal, and 95.6 on Long.
The consistent gains across suites corroborate that CV-Atten reduces cross-modal interference and stabilizes the learning of world-knowledge variations.
Overall, the ablation results validate the necessity of each component, and highlight that the joint use of PWKE and LWVQ is essential to fully realize the performance of $\boldsymbol{\Delta}$VLA.

\noindent\textbf{Ablation Study on the PWKE Module.}
Table~\ref{tb:ab_PKWE} examines how design choices in PWKE influence the overall performance of $\boldsymbol{\Delta}$VLA.
With only FiLM-guided region extraction, $\boldsymbol{\Delta}$VLA achieves 95.6 on Spatial, 97.2 on Object, 96.4 on Goal, and 93.4 on Long.
Adding semantic cues improves all suites to 96.8, 98.0, 96.8, and 93.8, suggesting better identification of instruction-relevant manipulable entities.
Incorporating depth cues further increases performance to 97.0, 98.2, 96.6, and 94.4, indicating the benefit of geometric priors under contact-rich interactions and long-horizon execution.
Enabling both semantic and depth cues yields the strongest results, which supports their complementarity.
We also evaluate instruction conditioning in region extraction.
Replacing FiLM-guided selection with an unguided alternative reduces performance to 97.4 on Spatial, 98.2 on Object, 96.2 on Goal, and 94.0 on Long, even with semantic and depth cues retained.
This degradation highlights the importance of textual modulation for aligning extracted regions with the instruction and avoiding task-irrelevant cues.

\noindent\textbf{Ablation Study on the Region and World Tokens in PWKE.} 
As shown in Table~\ref{tab:Region and World Tokens}, we investigate the effect of token configuration and attention design in PWKE.
Varying the number of the region and world tokens reveals a trade-off between spatial coverage and redundancy.
The configuration with 64 region tokens and 9 world tokens achieves the best overall performance, reaching 98.6 on Spatial, 99.4 on Object, 97.4 on Goal, and 95.6 on Long.
Increasing the number of world tokens beyond this setting provides no additional benefit and can slightly reduce accuracy, suggesting that excessive global tokens introduce redundant context and dilute decision-critical cues.
Likewise, increasing region tokens from 64 to 128 does not improve results, indicating that denser region sampling is unnecessary once manipulable areas are sufficiently represented.
We further compare attention modeling strategies.
Masked region-to-world attention consistently outperforms bidirectional attention, improving Spatial from 97.6 to 98.6 and Long from 94.4 to 95.6.
This supports the use of a structured information flow in which region tokens capture localized interaction evidence while world tokens aggregate global priors with reduced interference.
Overall, these results indicate that compact tokenization with constrained attention yields more stable interaction priors.

\noindent\textbf{Sensitivity to Pseudo-label Quality.}
To validate the robustness and practical reliability of PWKE under imperfect supervision, we inject controlled noise into the prior labels during training and report the results in Table~\ref{tab:Noisy Label}.
$\Delta$VLA remains highly stable under moderate noise.
With 30\% noise, it still achieves 98.0 on Spatial, 98.6 on Object, 96.6 on Goal, and 95.0 on Long, exhibiting only minor degradation from the clean-label setting.
A more apparent drop is observed only under severe corruption.
When 50\% noise is introduced, the performance decreases to 97.0, 97.4, 96.0, and 93.8 on the four suites, respectively.
Overall, these results indicate that PWKE does not rely on highly accurate pseudo-labels, and that the proposed framework can tolerate realistic levels of label noise, which is important when pseudo-labels are obtained from off-the-shelf pretrained models in real-world pipelines.

\noindent\textbf{Ablation Study on the LWVQ Module.} 
Table~\ref{tab:LWVQ} verifies that modeling action-induced variations is more effective than directly predicting future states.
When LWVQ is instantiated as full future modality reconstruction, the model achieves 95.6 on Spatial, 96.8 on Object, 95.6 on Goal, and 92.0 on Long.
Replacing future reconstruction with variation prediction consistently improves performance, reaching 96.8, 97.4, 96.4, and 93.2 with continuous variations.
The proposed latent variation design yields the best results.
These results indicate that predicting full future observations can be unnecessarily burdensome and may emphasize appearance-level details that are weakly coupled to control, whereas learning compact latent variations provides a more structured and decision-relevant target that better captures the world changes induced by actions.
We also examine how to parameterize variation tokens.
Using a shared set of variation tokens results in 96.8 on Spatial, 98.6 on Object, 96.0 on Goal, and 94.4 on Long.
Separating the variation tokens improves all suites, suggesting that disentangling semantic, geometric, and regional dynamics reduces interference and supports more accurate variation modeling.
Overall, the ablation demonstrates that the variation-based formulation is a key factor behind the performance gains, and that LWVQ further benefits from modality-aware token disentanglement.
\begin{table}[t]
\caption{\textbf{Ablation Study on the Number of Each Type of Variation Tokens.} We select a moderate number of variation tokens to achieve better performance.}
\vspace{-6pt}
\label{tab:Variation Tokens}
\centering
\footnotesize
\setlength{\tabcolsep}{12pt}
\begin{tabular}{c|cccc}
\toprule
\textbf{Method} & \textbf{Spatial} & \textbf{Object} & \textbf{Goal} & \textbf{Long} \\
\hline
\rowcolor[HTML]{FFF5CC}  
4 &\textbf{98.6}&\textbf{99.4}&\textbf{97.4} &\textbf{95.6}   \\
9 &98.4 &99.2 &97.0 &95.4 \\
16 &98.0 &98.8 &96.8 &95.0   \\
\bottomrule
\end{tabular}
\vspace{-5pt}
\end{table}
\begin{figure*}
        \centering
        \includegraphics[width=1\linewidth]{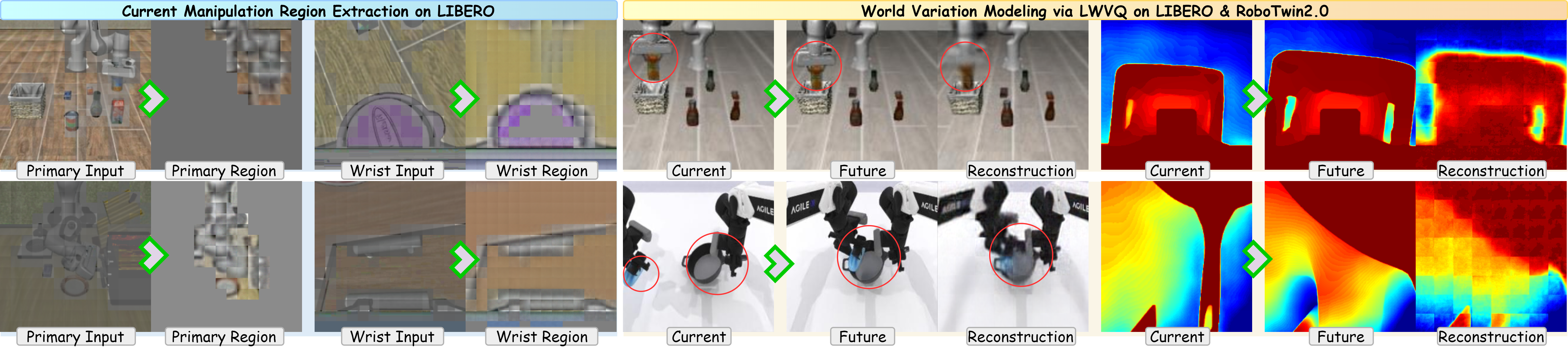}
\caption{\textbf{Visualization of Manipulation Region Extraction and World Variation Modeling.} 
On the left, the manipulable regions extracted by the PWKE module are shown. 
On the right, the future states, reconstructed from the current state and latent variation through the LWVQ module, closely align with the real future states and accurately highlight dynamic changes, marked by red circles.}
   
    \label{fig:vis_knowledge}
    \vspace{-5pt}
\end{figure*}

\noindent\textbf{Ablation Study on the Number of Variation Tokens.}
As shown in Table~\ref{tab:Variation Tokens}, we evaluate the impact of different numbers of variation tokens on model performance. Using a smaller number of tokens consistently yields the best success rates across all four LIBERO suites. These results suggest that a compact set of variation tokens is sufficient to capture decision-relevant world changes, whereas an overly large token set introduces redundancy and can dilute the learning signal, making variation modeling less focused and less stable.
\begin{figure*}
        \centering
        \includegraphics[width=1.0\linewidth]{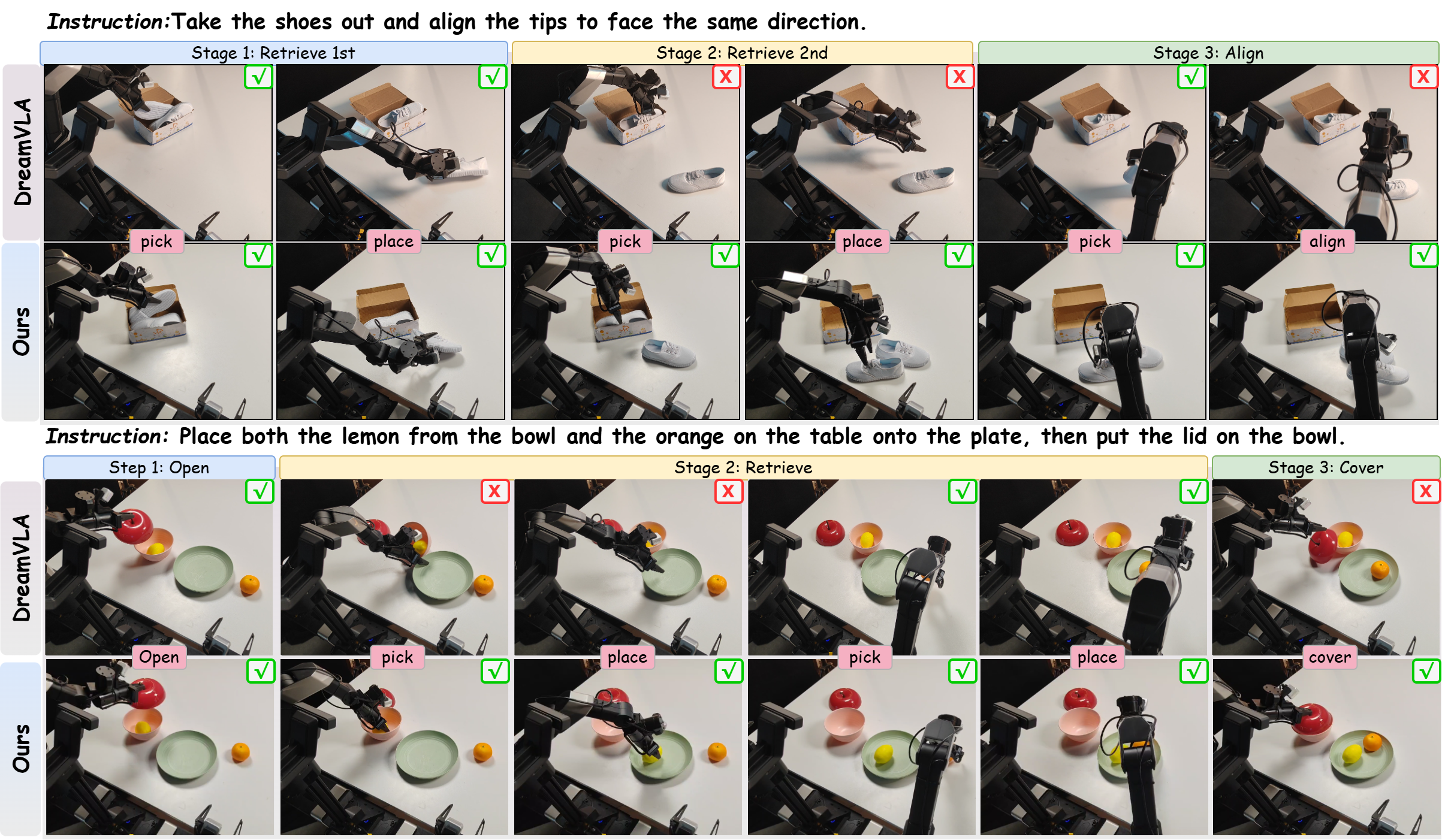}
\caption{\textbf{Visualization of Real-World Execution Comparison.} 
Given the long-horizon instruction, $\boldsymbol{\Delta}$VLA  successfully
 completes all stages of the manipulation task, while DreamVLA fails at multiple critical steps.}
   
    \label{fig:vis_real}
    \vspace{-5pt}
\end{figure*}
\begin{figure*}
        \centering
        \includegraphics[width=0.92\linewidth]{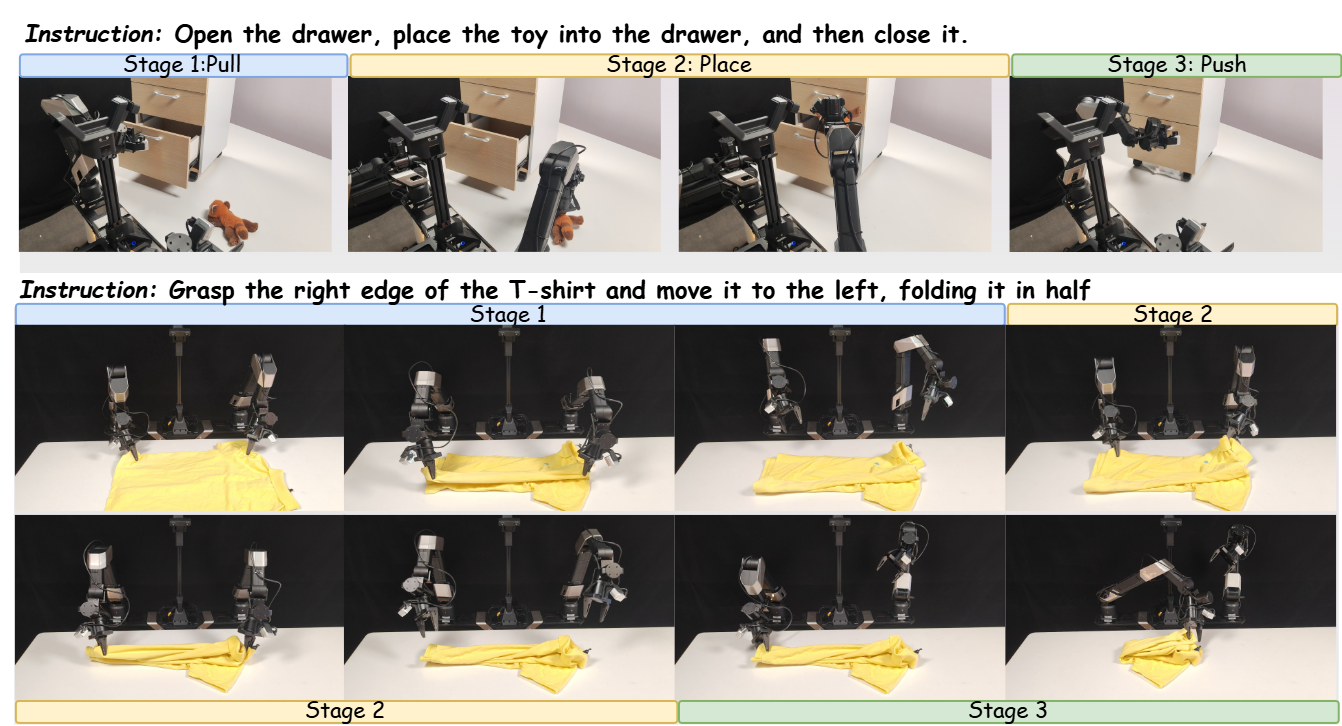}
   \caption{\textbf{Visualization of Executions on Two Additional Real-World Tasks.} $\Delta$VLA successfully completes all sub-stages in each task, demonstrating reliable long-horizon planning, robust manipulation skills, and accurate reasoning.}

    \label{fig:add_vis_real}
    \vspace{-12pt}
\end{figure*}
\begin{figure}
        \centering
        \includegraphics[width=1\linewidth]{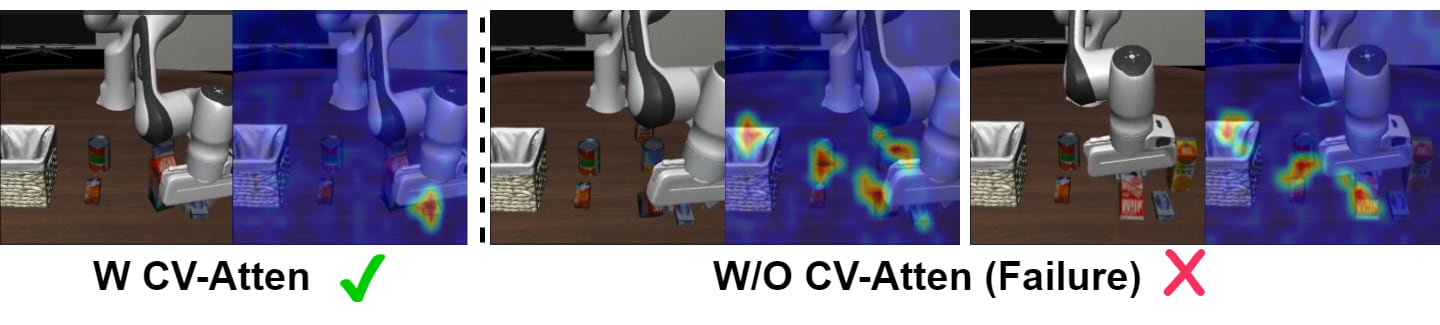}
\caption{\textbf{Diagnostic Analysis of CV-Atten.} During geometry-critical grasping, CV-Atten suppresses cross-modal interference by keeping attention on contact-relevant geometry. Without it, attention leaks to surrounding semantic cues, causing grasp-point drift and failure.}
   
    \label{fig:vis_atten}
    \vspace{-12pt}
\end{figure}

\subsection{Qualitative Analysis}
\noindent\textbf{Visualization of Manipulation Region Extraction and World Variation Modeling.}
As illustrated in Fig.~\ref{fig:vis_knowledge}, the left part visualizes the manipulable regions extracted by PWKE. The highlighted regions are well aligned with the contact-critical areas around the robot gripper and the target objects, indicating that the auxiliary-head supervision effectively drives the model to attend to task-relevant interaction cues rather than background clutter. This suggests that PWKE provides a reliable interaction prior that supports stable action generation under diverse scenes.
The right part visualizes the LWVQ reconstruction of future world states conditioned on the current observation and latent variations. The reconstructed futures closely match the ground-truth states, and more importantly, they faithfully capture the action-induced changes, such as object displacement and geometry transitions, as emphasized by the red circles. These qualitative results support our motivation that learning compact latent variations can preserve decision-critical dynamics while avoiding unnecessary appearance-level details, thereby improving controllability and long-horizon consistency.

\noindent\textbf{Visualization of Real-World Execution Comparison.}
Furthermore, Fig.~\ref{fig:vis_real} presents two representative real-world long-horizon comparisons under multi-step instructions.
In the first example, \emph{``Take the shoes out and align the tips to face the same direction,''} the task naturally decomposes into three stages, retrieving the first shoe, retrieving the second shoe, and finally aligning both shoes into a consistent orientation.
$\boldsymbol{\Delta}$VLA completes all stages with coherent stage-wise execution, successfully transitioning from the first retrieval to the second retrieval and then performing the final fine-grained alignment without losing the already-achieved progress.
In contrast, the predictive baseline DreamVLA exhibits failures at critical transition points.
It often becomes unstable when switching from the first retrieval to the second retrieval, and it frequently fails to execute the final alignment reliably, resulting in misaligned shoes or incomplete task completion.

In the second example, \emph{``Place both the lemon from the bowl and the orange on the table onto the plate, then put the lid on the bowl,''} the task requires three sequential stages, opening the bowl, retrieving and placing two different objects from distinct locations, and finally covering the bowl with the lid.
This setting is challenging because it involves repeated pick-and-place operations with different targets, along with a late-stage container closure that is sensitive to intermediate object placements.
$\boldsymbol{\Delta}$VLA follows the instruction order and completes the full sequence, successfully placing both the lemon and the orange onto the plate and finishing with a correct covering action.
By comparison, DreamVLA can complete the initial opening step, but it commonly breaks down during the retrieve-and-place stage, such as failing to pick the correct object, placing it outside the plate, or losing consistency when switching to the second object, which ultimately prevents successful completion of the final covering stage.

Across both cases, the predictive baseline failures concentrate on stage transitions, where small deviations in intermediate states lead to compounding errors and loss of task progress.
By explicitly modeling latent world variations, $\boldsymbol{\Delta}$VLA better tracks action-induced state changes and preserves decision-relevant cues for the next subgoal, leading to more stable long-horizon behavior and stronger real-world generalization under sequential object interactions and geometric constraints.

\noindent\textbf{Diagnostic Analysis of CV-Atten.}
To provide a diagnostic beyond success rates, we visualize cross-attention maps with and without CV-Atten in Fig.~\ref{fig:vis_atten}.
During grasping, reliable execution requires focusing on geometry-critical cues such as the object contour and contact-relevant structure, and the attention should concentrate on the target object rather than the surrounding scene.
With CV-Atten enabled, the attention is sharply localized around the gripper and the graspable contour of the target object, indicating geometry-aligned reasoning that supports accurate contact.
In contrast, removing CV-Atten induces cross-group leakage, where attention drifts toward semantically salient but geometrically less informative regions.
This semantic bias dilutes geometry-focused perception, leading to grasp-point drift, contact misalignment, and eventual failure.
Overall, the visualization suggests that CV-Atten suppresses cross-modal interference in geometry-critical interactions and stabilizes grasping by keeping attention anchored on the target object's geometry.

\noindent\textbf{Visualization of Executions on Two Additional Real-World Tasks.}
Fig.~\ref{fig:add_vis_real} visualizes executions of $\boldsymbol{\Delta}$VLA on two additional real-world long-horizon tasks that require sequential subgoals and precise contact-rich control.
In the drawer task with the instruction \emph{``Open the drawer, place the toy into the drawer, and then close it,''} the policy completes a three-stage sequence, pulling the drawer to a feasible opening, placing the toy into the drawer cavity, and finally pushing the drawer to close it.
This task is challenging because the interaction mode changes across stages, and each stage depends on the correct geometric state produced by the previous step.
$\boldsymbol{\Delta}$VLA maintains coherent progress across these transitions and completes the full instruction.

In the T-shirt folding task with the instruction \emph{``Grasp the right edge of the T-shirt and move it to the left, folding it in half,''} the execution involves sustained deformable-object manipulation with large appearance changes and sensitive geometric alignment.
As shown, $\boldsymbol{\Delta}$VLA performs stage-wise coordinated grasping and repositioning, preserves the intended fold direction, and completes the folding sequence with stable intermediate states.
It further corrects small misalignments by adjusting grasp points and repositioning trajectories as the cloth deforms, preventing error accumulation across stages.
Moreover, the intermediate folded configurations remain consistent with the intended geometric constraint (edge-to-edge alignment), facilitating a reliable transition to subsequent steps.
Together, these qualitative results demonstrate that $\boldsymbol{\Delta}$VLA can reliably track action-induced world changes over extended horizons, enabling consistent multi-step reasoning and robust execution under real-world sensing noise and actuation uncertainty.

\vspace{-5pt}
\section{Conclusion}
In this work, we presented $\boldsymbol{\Delta}$VLA, a variation-based predictive framework that models prior-grounded world-knowledge variations for robust and efficient robotic manipulation. The proposed \textbf{PWKE} module constructs an explicit current-world knowledge prior through auxiliary-head-guided extraction of manipulable, semantic, and geometric cues. The \textbf{LWVQ} module discretizes temporal variations via a learnable world-knowledge codebook, enabling compact and interpretable representation of dynamic changes. Moreover, the \textbf{CV-Atten} mechanism enforces disentangled causal reasoning by isolating modality-specific variations. Comprehensive experiments across simulation and real-world platforms demonstrate that $\boldsymbol{\Delta}$VLA achieves state-of-the-art performance, significantly improving both generalization and efficiency. 

\ifCLASSOPTIONcaptionsoff
  \newpage
\fi

\bibliographystyle{IEEEtran}
\bibliography{IEEEabrv,reference}

\end{document}